\documentclass{article} 
\usepackage{iclr2027_conference,times}

\usepackage{amsmath,amsfonts,bm}

\def\eqref#1{equation~\ref{#1}}

\def\1{\bm{1}}

\DeclareMathAlphabet{\mathsfit}{\encodingdefault}{\sfdefault}{m}{sl}
\SetMathAlphabet{\mathsfit}{bold}{\encodingdefault}{\sfdefault}{bx}{n}

\usepackage{hyperref}
\usepackage{url}
\usepackage{graphicx}
\usepackage{float}
\usepackage{placeins}
\usepackage{longtable}
\usepackage[textsize=scriptsize]{todonotes}
\usepackage{xcolor}
\usepackage[normalem]{ulem}
\usepackage{cleveref}
\usepackage{enumitem}
\creflabelformat{equation}{#2#1#3}  
\definecolor{chris}{RGB}{110,150,220}

\title{MuPACE: Learning to Pace Muon for LoRA Fine-Tuning}

\author{Jan Kobiolka, Christian Frey \& Josif Grabocka  \\
Department of Computer Science\\
University of Technology Nuremberg\\
Nuremberg, Germany \\
\texttt{\{jan.kobiolka, christian.frey,josif.grabocka\}@utn.de} \\
}

\newtheorem{definition}{Definition}[section]

\iclrfinalcopy

\begin{document}

\maketitle

\fancyhead[L]{Preprint}
\pagestyle{fancy}
\thispagestyle{fancy}

\begin{abstract}
Learning rate schedules are typically fixed before training begins, yet when fine-tuning large language models with low-rank adapters (LoRA), the appropriate scale varies substantially with the model, task, and adapter configuration. Practitioners re-tune a schedule for every new problem, yet a tuned schedule still cannot react to the training dynamics it encounters. We instead amortize learning rate selection across LoRA fine-tuning problems. We leave the base optimizer, Muon, untouched, so that its normalized update fixes the direction, and meta-learn only the global step size. Our method MuPACE is a small recurrent policy, meta-trained across a distribution of fine-tuning runs, that conditions on the observed trajectory and emits bounded adjustments to an interpretable analytic schedule. On 64 fine-tuning problems, MuPACE outperforms standard fixed schedules at a strong shared default on 64–78\% of problems and is on par with the same schedules with per-task tuned peaks. Meta-trained only on models with up to 16 layers, MuPACE transfers one-shot to unseen tasks on models as deep as 64 layers, such as Qwen3-32B and Qwen3.8-27B, where a single run remains competitive with swept schedules.


\end{abstract}
\section{Introduction}
\label{sec:intro}

Modern deep learning optimizers increasingly decouple the update direction from its overall scale. Adaptive methods such as Adam \citep{Kingma_2014_adam} and AdamW \citep{loshchilov2018decoupled} rescale gradients coordinate-wise, while more recent matrix-aware optimizers such as Shampoo \citep{gupta2018shampoo}, SOAP \citep{vyas2025soap}, and Muon \citep{jordan2024muon} exploit the structure of neural network parameters. Muon, in particular, constructs updates for matrix-valued parameters by orthogonalizing the momentum update and has shown promising performance in language-model pre-training \citep{liu2025muonscalablellmtraining} and LoRA \citep{hu2022lora} fine-tuning \citep{qu2026can}. 

The perspective of separating the update direction from the step size is also reflected in the norm-based view of steepest descent \citep{bernstein_newhouse_2024}. At iteration $t$, the optimizer forms an optimization signal $s_t$ from the gradient $g_t = \nabla_\theta \ell(\theta_t)$, such as the gradient itself or the momentum estimate. We denote by $\mathcal{O}$ the optimizer-specific map from the optimization
signal $s_t \in \mathbb{R}^{m\times n}$ of an $m\times n$ matrix to an
update direction $d_t \in \mathbb{R}^{m\times n}$ of the same shape, with the
update scaled by the learning rate $\eta_t \in \mathbb{R}^+$, i.e.: 
\[
d_t=\mathcal{O}(s_t),
\qquad
d_t\in\arg\max_{\|d_t\|\leq 1}\langle s_t,d_t\rangle,
\qquad
\theta_{t+1}=\theta_t-\eta_t d_t.
\]
The norm determines the geometry of $d_t$, while the step size $\eta_t$ sets its magnitude. For example, the Euclidean norm yields the normalized direction $s_t/\|s_t\|_2$, whereas the infinity norm yields the element-wise sign direction. For a matrix-valued signal $s_t = U_t\Sigma_tV_t^\top$, Muon uses the corresponding spectral-norm direction $d_t = U_tV_t^\top$, approximated in practice with a small number of Newton--Schulz iterations \citep{jordan2024muon}.

For normalized optimizers, the learning rate therefore plays a particularly important role. Normalization can improve progress in challenging landscapes such as river valleys \citep{shen2026towards}, but it also reduces the influence of gradient magnitude on the update size. A small residual gradient can still be mapped to a direction of approximately fixed norm, leaving \(\eta_t\) responsible for determining whether the optimizer makes useful progress or overshoots the local solution. Step-size selection is therefore an important and distinct control problem.

In practice, this control problem is often solved by committing to a learning rate schedule before training begins. Common choices such as cosine decay \citep{loshchilov2017sgdr} and warmup-stable-decay (WSD) \citep{hu2024minicpm} set \(\eta_t\) primarily as a function of training progress, while their overall scale is selected through hyperparameter optimization. This is expensive because finding a good learning rate schedule can require several training runs, and the resulting setting does not transfer across fine-tuning problems, as the appropriate schedule varies with the downstream task, adapter parameterization, batch size, and optimization budget \citep{biderman2024lora, lee2026beware}.  This lack of transfer makes fixed schedules especially costly, since tuning the peak learning rate for every new problem adds an extra search procedure, while the resulting schedule still cannot react to the optimization trajectory.

We therefore ask the following question: \textit{Can learning rate schedule selection instead be amortized across LoRA fine-tuning problems?} We introduce \textbf{MuPACE} (\textbf{Mu}on \textbf{P}olicy for \textbf{A}daptive \textbf{C}ontrol across LoRA Fine-tuning \textbf{E}pisodes), a learned, trajectory-conditioned learning rate scheduler for LoRA. LoRA fine-tuning is especially relevant because only a few large organizations train large language models from scratch, while most users adapt existing pretrained models \citep{bommasani2021opportunities}. Unlike previous learned optimizers that meta-learn the update rule itself \citep{metz2022velo,moudgil2025celo}, we leave Muon unchanged and meta-learn only its global step-size schedule. A small recurrent policy observes the optimization trajectory and predicts bounded modifications to an analytic schedule family, explicitly separating changes in overall scale from changes in schedule shape. The scheduler is meta-trained across a distribution of fine-tuning episodes and then applied one-shot to new problems, producing an adaptive schedule within a single training run without hyperparameter search or downstream-specific tuning.

To assess the effectiveness of our approach, we evaluate whether a single meta-trained scheduler can replace per-task learning rate search across a broad range of fine-tuning settings, spanning pretrained models, downstream tasks, adapter configurations, batch sizes, and training horizons. On 64 problems, MuPACE outperforms strong fixed defaults on 64--78\% of tasks and performs on par with schedules tuned separately for each problem, while requiring only a single trial. Moreover, despite being meta-trained only on models with up to 16 layers, MuPACE generalizes in a one-shot manner to large-scale models such as Qwen3-32B and Qwen3.8-27B (both with 64 layers) on unseen downstream tasks. It achieves competitive performance close to that of swept schedules while requiring only a single run.

Our key contributions are:
\begin{itemize}[leftmargin=*]
    \item We introduce \textbf{MuPACE}, a meta-learned, trajectory-conditioned learning rate scheduler for LoRA fine-tuning that adapts a bounded analytic schedule while leaving the base optimizer, Muon, unchanged, enabling one-shot application without target-task learning rate search.
    
    \item We show that a single meta-trained scheduler generalizes across fine-tuning problems and model scales. On 64 problems, MuPACE outperforms strong fixed defaults and performs comparably to per-task tuned schedules, while transferring one-shot to Qwen3-32B and Qwen3.8-27B on unseen tasks, where it remains competitive with tuned learning rate schedules in a single run.
\end{itemize}

\section{Related Work}
\paragraph{Learning rate scaling and schedules.}
The maximal update parameterization ($\mu$P) jointly defines width-dependent initialization, parameter multipliers, and per-layer learning rates, enabling near-optimal hyperparameters to transfer across width \citep{Tensor_Programs_V}. Depth-$\mu$P extends this principle to residual networks with single-layer blocks \citep{Tensor_Programs_VI}. Transfer has been related to approximately width- and depth-independent Hessian sharpness under feature learning \citep{noci2024super}, though for Adam-trained Transformer language models, standard parameterization with suitable per-layer learning rates also transfers and can outperform $\mu$P when per-layer constants are tuned \citep{Everett_Scaling}. A parallel line of work concerns the shape of the schedule rather than its scale. Cosine annealing \citep{loshchilov2017sgdr} remains common in LLM pretraining, alongside warmup-stable-decay (WSD) schedules that maintain a constant learning rate before a terminal cooldown \citep{hu2024minicpm}. Additional approaches include linear decay to zero \citep{bergsma2025straight} and schedule-free optimization \citep{defazio2024road}. However, frontier labs remain divided on the choice of learning rate schedule. Kimi K2 \citep{team2025kimi2} uses WSD, DeepSeek V4 \citep{xu2026deepseek4} uses a warmup, long constant plateau, and terminal cosine cooldown, whereas GLM-4.5 \citep{zeng2025glm} and Kimi K3 \citep{team2026kimi3} report favoring cosine over WSD. Fixed-horizon theory similarly predicts power decay in easier regimes and a WSD-like schedule in harder ones \citep{li2026optimal}. Much of this work, however, concerns pretraining from random initialization over a known token budget, and does not establish that either the learning rate scale or the schedule shape transfers across LoRA fine-tuning problems.

\paragraph{Online learning rate adaptation.}
Learning rates can also be adapted during training using heuristic or learned rules, rather than prescribed by a fixed schedule in advance. For example, the heuristic ReduceLROnPlateau scheduler \citep{lronplateau} reduces the learning rate when a monitored metric stops improving. More flexible schedules can be learned by differentiating through the training trajectory \citep{pmlr-v37-maclaurin15}, while Hypergradient Descent adapts learning rates online using hypergradients \citep{gunes2018online}. Related approaches include Self-Tuning Networks for bilevel online hyperparameter optimization \citep{mackay2018selftuning}, D-Adaptation for automatically estimating an appropriate learning rate scale \citep{defazio2023learning}, and MetaOptimize for adapting step sizes based on their longer-term effects on future losses \citep{sharifnassab2025metaoptimize}. In contrast, we amortize learning rate adaptation across a distribution of fine-tuning problems. After meta-training, a frozen trajectory-conditioned policy adapts one-shot to new problems within a single fine-tuning run, without downstream specific hyperparameter search.

\paragraph{Learned optimizers and schedulers.}
Learning-to-learn replaces hand-designed update rules with meta-learned ones \citep{andrychowicz2016learning,metz2022practical}. VeLO \citep{metz2022velo} scales this methodology to a general-purpose optimizer using a hierarchical architecture in which a per-tensor LSTM hypernetwork generates the weights of a small per-parameter MLP update rule. It requires no per-task optimizer hyperparameter tuning but incurs roughly 4,000 TPU months of meta-training. Celo \citep{moudgil2025celo} obtains strong out-of-distribution meta-generalization with only 24 GPU hours by combining task augmentation with a learned per-parameter MLP update rule and a learned global scheduler, meta-trained in two stages. Celo2 \citep{moudgil2026celo2} further separates update-rule learning from step-size selection. It meta-learns only the update rule, leaves the step size tunable, and defers learning a generally applicable scheduler to future work. Closest to our setting, L3RS specializes learned optimization to fine-tuning pretrained models and learns, for each layer, both an update magnitude and a mixture of directions supplied by base optimizers using temporal and optimization-statistic features \citep{kristiansen2024narrowing}. Our approach instead focuses on a smaller part of the optimization process. We keep the update direction from a fixed base optimizer, Muon, and meta-learn only the global learning-rate schedule for low-rank fine-tuning.
\section{Methodology}
\label{sec:method}

\subsection{Problem Setting}
\label{sec:task}

We consider low-rank fine-tuning problems of the form $\theta^\star \in \arg\min_{\theta}\ell_{\mathrm{tr}}(\theta;\mathcal{T}),
\mathcal{T} \sim p(\mathcal{T})$, 
where $\theta$ denotes the trainable adapter factors of a frozen pretrained model, $\ell_{\mathrm{tr}}$ is the training loss, and $p(\mathcal{T})$ is a distribution over fine-tuning task instances.

\begin{definition}[Fine-tuning Task Instance]
\label{def:task}
A task instance $\mathcal{T} = (m, r, \mathcal{D}, a, T, b, \xi)$ specifies
adaptation of a pretrained model $m$ by replacing each target weight
$W \in \mathbb{R}^{d^{\mathrm{out}} \times d^{\mathrm{in}}}$ with
\begin{equation}
W + \tfrac{a}{r}\, B A,
\qquad B \in \mathbb{R}^{d^{\mathrm{out}} \times r},\;
A \in \mathbb{R}^{r \times d^{\mathrm{in}}},
\qquad r \ll \min(d^{\mathrm{out}}, d^{\mathrm{in}}).
\end{equation}
Only the adapter factors $\theta = \{(A, B)\}$ are trained on a dataset
$\mathcal{D}$, with splits $(\mathcal{D}^{\mathrm{tr}},\mathcal{D}^{\mathrm{val}})$, for $T$ steps at batch size $b$. The seed $\xi$ fixes initialization and data
order. One such run is an \emph{episode}.
\end{definition}

The training loss is the mean next-token cross-entropy over target tokens in
$\mathcal{D}^{\mathrm{tr}}$. We record the episode's trajectory after $t$ steps as
\begin{equation}
\tau(t) = \big\{\,(\ell_i,\; \ell^{\mathrm{val}}_i,\; g_i,\; i/T)\,\big\}_{i=1}^{t},
\label{eq:traj}
\end{equation}
where $\ell_i$ and $\ell^{\mathrm{val}}_i$ are mini-batch training and validation
losses, $g_i$ is the adapter gradient of the training loss, and $i/T$ is
normalized progress. Validation batches come from $\mathcal{D}^{\mathrm{val}}$. Their losses are provided to the scheduler as trajectory information, but their gradients are not computed or used for optimization.

Both the learning rate scale and schedule shape can depend on the model, sub-task, adapter, and training configuration. Standard schedules, however, are fixed functions of $t/T$ and ignore the training dynamics they are applied to. We instead propose to learn a policy with meta-parameters $\phi$,
\begin{equation}
\eta_t = \pi_\phi(z_t),
\end{equation}
where $z_t$ summarizes static task descriptors and training statistics $\tau(t)$ available
before the update at step $t$ (Section~\ref{sec:meta-model}). The policy learns only the scalar step size. The underlying optimizer, Muon, determines the update direction. We meta-train the policy across task instances to minimize a validation-loss objective within each episode's budget (Section~\ref{sec:outer}).

\subsection{Inner Optimization}
\label{sec:inner}

\paragraph{Optimization Step.}
We keep the base weights frozen, so every trainable parameter of a
task instance is a LoRA factor $A$ or $B$ and hence matrix-shaped. We can
therefore update the entire adapter with Muon and require no element-wise
fallback rule. The inner update is $\theta_{t+1} = \theta_t - \eta_t d_t$, with
$d_t = \mathcal{O}(s_t)$ denoting the Muon direction. We post-scale each Muon update to unit RMS \citep{moudgil2026celo2} before applying the learning rate, compensating for its RMS decreasing with matrix size \citep{liu2025muonscalablellmtraining}.

\begin{figure}[t]
\centering
\includegraphics[width=\linewidth]{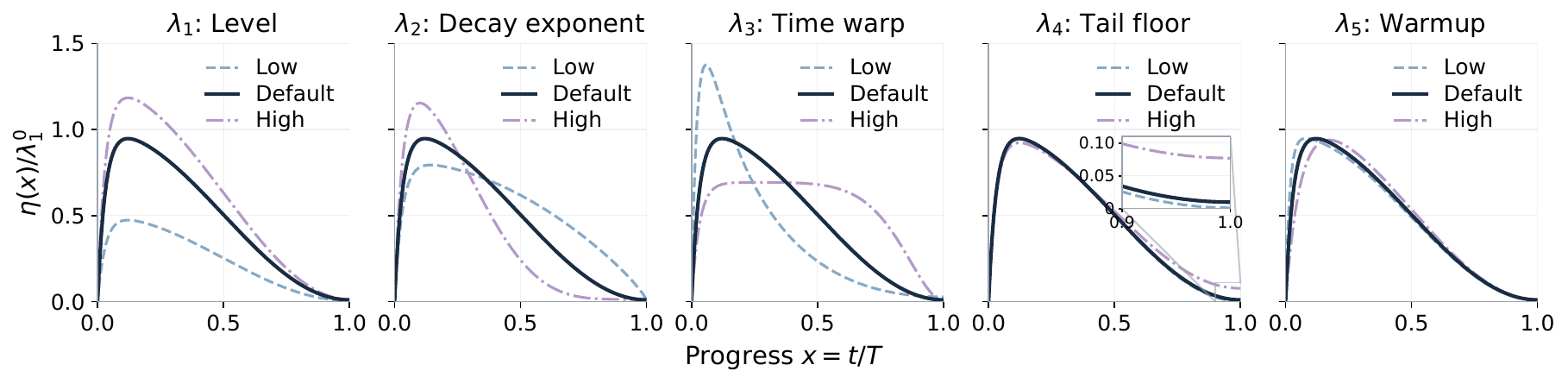}
\vspace{-14pt}
\caption{Shape-normalized schedules varying one parameter at a time, default in dark.}
\label{fig:schedule-family-parameters}
\end{figure}

\paragraph{Schedule Family.}
We do not emit $\eta_t$ as a free function of the trajectory. Instead, we
constrain the policy to a hand-designed analytic family and let it emit bounded
deviations from it. For normalized progress $x = t/T$,
\begin{align} 
\eta(x) &= \lambda_1 \cdot s(x), \label{eq:sched} \\
s(x) &= \mathrm{warmup}(x;\lambda_5) \cdot \big[(1-\lambda_4)\,\mathrm{decay}(x;\lambda_2,\lambda_3) + \lambda_4\big], \nonumber \\
\mathrm{warmup}(x;\lambda_5) &= 1 - \exp(-x/\lambda_5), \qquad
\mathrm{decay}(x;\lambda_2,\lambda_3) = \mathrm{clip}\!\left(\cos\!\left(\tfrac{\pi}{2}x^{\lambda_3}\right), \delta, 1\right)^{\lambda_2}. \nonumber
\end{align}
The family is controlled by a 5-dimensional vector $\lambda = (\lambda_1, \ldots, \lambda_5)$, which is what the policy emits. The parameters split into a scale and a shape. $\lambda_1$ sets the overall scale of the schedule, while the other four parameters determine its shape. Specifically, $\lambda_2$ controls the decay rate, $\lambda_3$ shifts the knee in time, $\lambda_4$ sets the tail floor, and $\lambda_5$ controls the warmup timescale.
\Cref{fig:schedule-family-parameters} illustrates the effect of each parameter. The policy re-emits
$\lambda_1, \ldots, \lambda_4$ at every inner step, and the resulting $\eta_t$
is the global learning rate of the adapter. The warmup constant $\lambda_5$ is emitted once, at the start of the episode, and held fixed for its remainder. This analytic form combines gradual warmup with flexible cosine decay, keeping the policy’s output space low-dimensional and interpretable while allowing bounded adjustments to a known-good default.

\paragraph{Shape Normalization.}
 Changing the shape parameters also changes the schedule’s overall magnitude. Meta-training could therefore exploit the shape parameters to adjust magnitude rather than the shape. To separate shape from scale, we normalize each curve to the default shape’s RMS. 
\begin{equation} 
\eta(x) = \lambda_1 \cdot \frac{\lVert s_0 \rVert}{\lVert s \rVert}\, s(x),
\qquad
\lVert s \rVert = \Big(\textstyle\int_0^1 s(u)^2\,\mathrm{d}u\Big)^{1/2},
\label{eq:norm}
\end{equation}
where $s_0$ is the default shape. At each step, we approximate the norm using
65 uniformly spaced points on $[0,1]$. This leaves the default schedule
unchanged and lets $\lambda_1$ control each curve's RMS scale. The normalization
applies to the curve defined by the current parameters, not to the full
trajectory as those parameters change across steps.

\paragraph{Depth Scaling.}
The emitted step size is finally rescaled by a fixed factor depending only on
the depth $L_{\mathrm{d}}$, i.e.\ the number of layers, of the base model,
\begin{equation}\label{eq:depth}
\eta_t \;\leftarrow\; \left(\frac{L_{\mathrm{ref}}}{L_{\mathrm{d}}}\right)^{\!1/2} \eta_t,
\end{equation}
following Depth-$\mu$P \citep{Tensor_Programs_VI}, which prescribes
$\eta \propto 1/\sqrt{L_{\mathrm{d}}}$ for a scale-invariant update. We set
$L_{\mathrm{ref}}$ to the geometric center of the meta-training depths, so the
factor is $1$ in the middle of that range and shrinks the step size for deeper
models. 

\subsection{Meta-Model}
\label{sec:meta-model}

We parameterize $\pi_\phi$ as a small LSTM $f_\phi$ with one linear head per
schedule parameter, evaluated once per inner step so that it can condition on
the episode so far. Each raw output $h_{i,t}$ becomes a bounded multiplicative
deviation from a hand-picked default $\lambda_i^0$,
\begin{equation}
\lambda_{i,t} = \lambda_i^{0} \cdot \exp\!\left(c_i \tanh h_{i,t}\right),
\qquad i = 1, \ldots, 5,
\label{eq:bound}
\end{equation}
which keeps $\lambda_{i,t}$  within $\left[\lambda^0_i e^{-c_i},\, \lambda^0_i e^{c_i}\right]$. The hyperparameter $c_i$ sets the width of this range. We compute $\lambda_5$ once from the static features and $\log T$ before the first step and update the other four parameters at every step,
as discussed in \Cref{sec:inner}. The output layer is zero-initialized, so meta-training starts
exactly from the default schedule.

\paragraph{Features.}
We feed $f_\phi$ a $20$-dimensional feature vector in three groups.
\emph{Time} (4) contains $[x, \sin \pi x, \cos \pi x]$, where $x=t/T$, providing a smooth low-frequency encoding of position within the episode. We also include the total budget $\log T$.
\emph{Static} (5) is fixed throughout an episode. It contains the base-model width $\log W$ and layer count $\mathrm{depth}$, the factor $\log(a/r)$ by which the adapter enters the model, and the individual rank $\log r$ and scaling $\log a$.
\emph{Dynamic} (11) is recomputed at every step to summarize $\tau(t)$. It includes the initial loss level, the training loss relative to that initial value, and its local slope measured by the ratio of a fast to a slow EMA. For validation loss, we include the ratio of its fast EMA to its initial loss and the ratio of its fast EMA to its slow EMA. We also include the ratio of validation to training fast EMAs as an overfitting signal. The initial validation loss is used only for normalization and is not a separate feature. The remaining features are the cosine similarity between the current gradient and momentum, a gradient signal-to-noise proxy based on the bias-corrected momentum and second-moment accumulators, parameter-RMS growth relative to initialization, and the log parameter- and gradient-RMS of the adapter. 



\subsection{Meta-Training}
\label{sec:outer}

\paragraph{Scale Augmentation.}
Inferring the adapter scale from the trajectory only pays off if meta-training
varies it. We therefore reparameterize every sampled task instance. A
scalar $\kappa$ is drawn log-uniformly, the adapter factors are initialized at $1/\kappa$
times their usual value, and the forward pass multiplies them back by $\kappa$ \citep{moudgil2025celo}. This
leaves the function at step $t=0$ and the optimization problem unchanged, but not
the parameter RMS, the gradient RMS, or the step size that suits them. The static features alone are therefore no longer enough to determine the right step size, and the policy has to adjust it as the scale becomes visible in the trajectory.

\paragraph{Objective.}
We train the policy parameters $\phi$ to minimize the expected validation loss
across a distribution of fine-tuning tasks 
\begin{equation}
\min_{\phi}\; \mathbb{E}_{\mathcal{T} \sim p(\mathcal{T})}\,
\mathbb{E}_{\tau \sim \pi_\phi(\mathcal{T})}\,
\mathcal{L}_{\mathrm{final}}(\tau(T)).
\label{eq:meta-obj}
\end{equation}
For each episode, we skip the first $T_0 = \lfloor fT \rfloor$ steps and average
the remaining losses:
\begin{equation}
\mathcal{L}_{\mathrm{final}}(\tau(T)) = \frac{1}{T - T_0}\sum_{t = T_0 + 1}^{T} \tilde{\ell}^{\mathrm{val}}_t,
\label{eq:obj}
\end{equation}
Here $f \in [0,1)$ is the fraction of steps we skip, and
$\tilde{\ell}^{\mathrm{val}}_t$ is the log validation loss smoothed with an
exponential moving average. Taking the log focuses on relative improvements, while skipping the early steps ensures the score reflects the model’s performance during the later stages of training, rather than being dominated by the large initial loss. The policy observes validation loss, but inner updates use only training-loss
gradients. 

\paragraph{Meta-Gradient.}
We estimate the meta-gradient with truncated PES \citep{vicol2021unbiased}.
Each outer iteration unrolls an antithetic pair $\phi \pm \varepsilon$, $\varepsilon \sim \mathcal{N}(0, \sigma^2 I)$, for $L$
inner steps on $N$ task instances in parallel, and each instance is resampled
once it reaches its budget $T$. Since the meta-loss has to be emitted per inner
step, we emit $\tilde{\ell}^{\mathrm{val}}_t$ masked to $t > T_0$ and rescaled
by $T/(T - T_0)$. Averaged over the episode, this recovers
Equation~\ref{eq:obj}.

\section{Experimental Protocol}
\label{sec:protocol}

We meta-train MuPACE on a distribution over
LoRA fine-tuning tasks built from a deck of $37$ pretrained nanochat models \citep{nanochat} spanning depths $\{8, 12, 16\}$ and widths $128$ to $2048$, trained on $100$ Lots-of-LoRAs sub-tasks \citep{wang-etal-2022-super}, ranks
$r \in \{8, 16, 32\}$ and batch sizes $b \in \{8, 16, 32\}$. Conditioned on such a cell, the adapter scaling is drawn as $a \sim \mathrm{LogUniform}[8, 32]$, the
budget as $T \sim \mathrm{LogUniform}[100, 300]$, and the seed $\xi$. We place adapters on the query and value ($Q$ and $V$) projections of every attention block \citep{hu2022lora}. We initialize $A$ with a random Gaussian distribution and $B = 0$, so training starts from the pretrained model, and scale the update $BA$ by $a/r$ \citep{hu2022lora}. MuPACE therefore must handle combinations of model scale, adapter capacity and scaling, downstream task, and training length. Full parameter ranges, the deck listing, and the sub-task ids are given in Appendix~\ref{app:prior}. We open-source our implementation to foster future research\footnote{\url{https://anonymous.4open.science/r/MUPACE-3F54/README.md}}.

\subsection{Architecture and Training}
\label{sec:arch-training}

MuPACE is an LSTM with a hidden width of $48$ with one linear head per schedule parameter, ${\sim}20$K meta-parameters in total. We estimate the meta-gradient with PES at $\sigma = 0.01$ over a truncation window of $L = 50$ inner steps, keeping $N = 8$ task instances in flight and averaging the estimate over them. We train on the objective in Equation~\ref{eq:obj} with $f=0.2$, scoring the smoothed log validation loss over the last $80\%$ of each episode to emphasize consistent performance beyond the initial loss drop. An EMA with decay $0.9$ reduces batch noise. Each task runs for its budget $T$ and then is resampled. The scale augmentation of Section~\ref{sec:outer} draws $\kappa \sim \mathrm{LogUniform}[0.05, 20]$. For the depth factor of Equation~\ref{eq:depth} we set $L_{\mathrm{ref}} = \sqrt{8 \cdot 16} \approx 11.31$, the geometric centre of the meta-training depths. Meta-parameters are updated with AdamW at an outer learning rate of $10^{-3}$ on a cosine decay with gradient clipping. We train for $9000$ outer iterations, which takes roughly $72$ hours on four NVIDIA H100 GPUs. Remaining hyperparameters are reported in Appendix~\ref{app:hyperparams}.

\subsection{Baselines}
\label{sec:baselines}

We compare MuPACE against three fixed learning rate schedules, \emph{cosine} \citep{loshchilov2017sgdr}, \emph{constant} (warmup to a flat peak) and \emph{WSD} \citep{hu2024minicpm}, which we run at a default peak of $10^{-3}$, matching MuPACE's default schedule scale for $\lambda^0_1$ (H1 in \Cref{H3}) and with the peak tuned per task by validation loss (H2 in \Cref{H3}). In the out-of-distribution setting (H3 in \Cref{H3}) we add \emph{Hypergradient Descent} \citep{gunes2018online}, which adapts the learning rate online from the dot product of consecutive gradients and, like MuPACE, needs no per-task sweep. We run it as published and with the depth-$\mu$P post-scaling of Equation~\ref{eq:depth} applied to its learning rate, so that it receives the same multiplier as our learned scheduler. Every method runs the same Muon update from the same adapter initialization, so only the learning rate over time differs.

\subsection{Benchmarks}
\label{sec:benchmarks}

\paragraph{In-distribution performance.}
We evaluate MuPACE on $64$ configurations sampled from the meta-training distribution with scale augmentation disabled, each fixing the model, dataset, rank, scaling, batch size, and budget. All methods share the initialization and data order. The fixed schedules run either at their default peak learning rate of $10^{-3}$, which is at or near the best fixed rate for all three families (H1), or at a peak selected per task from seven log-spaced values in $[10^{-4}, 10^{-2}]$ (H2). MuPACE uses a single run. We report test loss at the iteration with the lowest validation loss.

\paragraph{Out-of-distribution transfer.}
We transfer MuPACE one-shot to \textbf{Qwen3-32B} \citep{yang2025qwen3} and \textbf{Qwen3.8-27B} \citep{qwen38}. Both have $64$ layers and width $5120$, against at most $16$ layers and width $2048$ in the meta-training deck. Further, Qwen3.8-27B adds a hybrid Gated DeltaNet linear-attention architecture the policy never saw. Adapters use $r = 8$, $a = 16$ and $T = 200$ steps on eight unseen datasets spanning instruction following, math, code, summarization, question answering and translation (\Cref{app:ood_datasets}). We report test loss at the iteration with the lowest validation loss. For fixed-schedule baselines, we additionally select the peak learning rate from five log-spaced values in $[10^{-5}, 10^{-3}]$, while MuPACE and each Hypergradient Descent variant use a single run.

\section{Experiments}
\label{sec:experiments}

\begin{figure}[t]
  \centering
  \includegraphics[width=\linewidth]{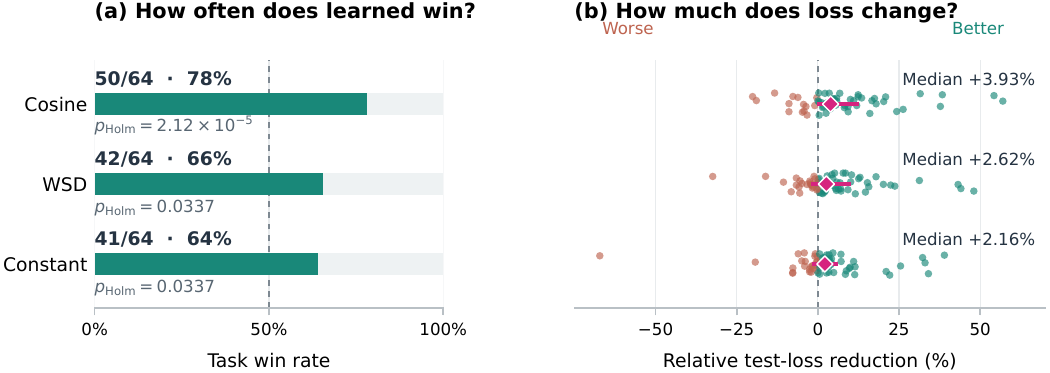}
  \vspace{-8pt}
  \caption{Learned scheduler versus defaults on $64$ tasks.
\textbf{(a)} Win rates and Holm-adjusted sign-test $p$ values.
\textbf{(b)} Relative test-loss reduction versus each baseline, where positive favors learned. Dots show tasks, diamonds medians, and bars interquartile ranges.}
  \label{fig:h1-summary}
\end{figure}

\paragraph{H1: Meta-learned schedulers can outperform standard default learning rate schedules on in-distribution fine-tuning problems.}\mbox{}\\
By default, MuPACE learns a bounded correction to a cosine learning-rate schedule, where the base schedule is scaled to have a peak learning rate of $10^{-3}$. H1 therefore tests the value of this learned correction against cosine, WSD, and constant schedules at that same fixed peak. Win rates are tested against 50\% using two-sided exact paired sign tests \citep{dixon1946statistical} with Holm correction \citep{Holm1979ASS} across the three baselines ($\alpha = 0.05$).

\begin{figure}[t]
  \centering
  \begin{minipage}[t]{0.49\linewidth}
    \vspace{0pt}
    \centering
    \includegraphics[width=0.8\linewidth]{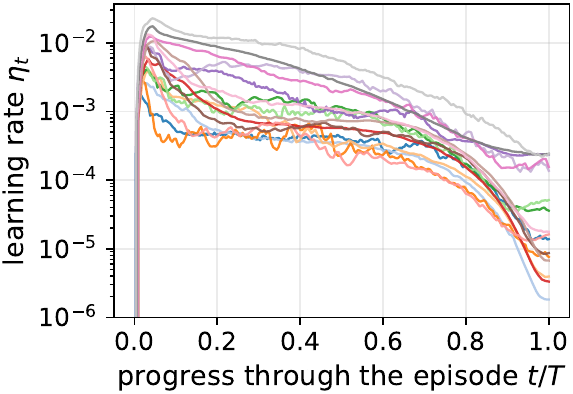}
    \vspace{-10pt}
    \caption{Learning rate applied by the learned scheduler on $16$ of the $64$ in-distribution tasks.}
    \label{fig:learned-schedules}
  \end{minipage}\hfill
  \begin{minipage}[t]{0.49\linewidth}
    \vspace{0pt}
    \centering
    \includegraphics[width=\linewidth]{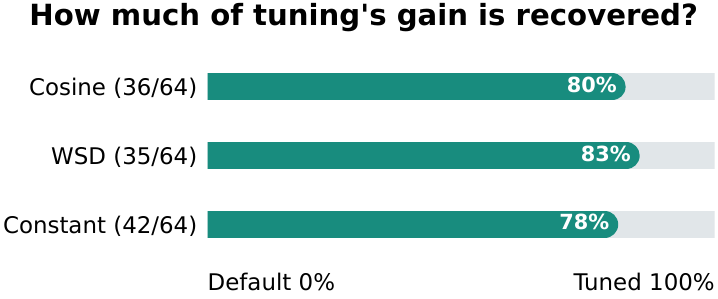}
    \vspace{-10pt}
    \caption{Median fraction of the per-task tuning gain recovered by the learned scheduler in one run, on the draws where tuning improves test loss by at least $0.5\%$.}
    \label{fig:h2-recovery}
  \end{minipage}
\end{figure}

The learned scheduler achieves lower test loss on 78\%, 66\%, and 64\% of draws against cosine, WSD, and constant, respectively, all significant after correction (Figure \ref{fig:h1-summary}). Median improvements are 2--4\% across the three comparisons. A further check against cosine at its best global scale ($2.154 \times 10^{-3}$)\footnote{39 of 64 draws, sign test p = 0.10.} gives a smaller, non-significant win-rate advantage, while the learned scheduler retains a median reduction of about 4\%. Overall, the learned correction provides gains beyond a strong fixed default by adapting the schedule to the task and to the observed training trajectory (Figure~\ref{fig:learned-schedules}).

\paragraph{H2: Meta-learned schedulers can eliminate per-task learning rate search while retaining competitive performance on in-distribution fine-tuning problems.}\mbox{}\\
We repeat the H1 comparison against the per-task tuned baselines, comparing their validation-based learning rate search with our learned scheduler, which requires no per-task sweep. Sign tests are applied as in H1.

\begin{figure}[t]
  \centering
  \includegraphics[width=\linewidth]{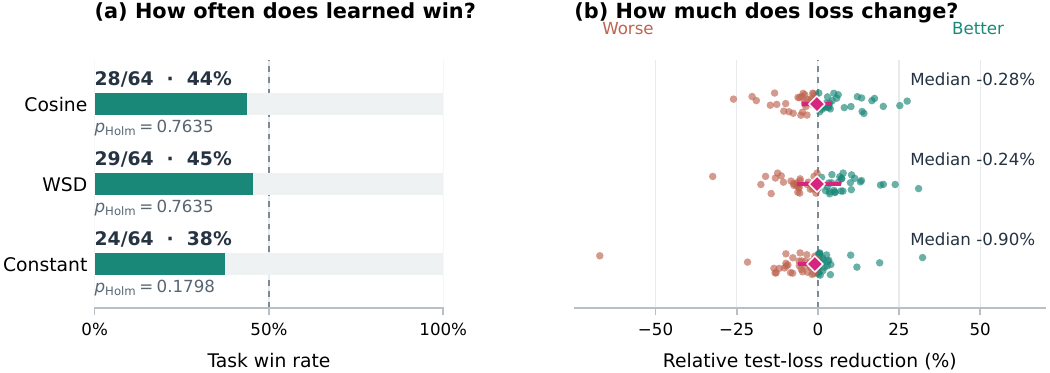}
  \vspace{-8pt}
  \caption{Learned scheduler versus validation-tuned baselines on $64$ tasks.
\textbf{(a)} Win rates and Holm-adjusted sign-test $p$ values.
\textbf{(b)} Relative test-loss reduction versus each tuned baseline, where positive favors learned. Dots show tasks, diamonds medians, and bars interquartile ranges.}
  \label{fig:h2-summary}
\end{figure}

Figure~\ref{fig:h2-summary} shows that the learned scheduler approaches the performance of per-task tuned schedules without the cost of a learning rate sweep. Our learned scheduler wins on $44\%$, $45\%$, and $38\%$ of draws
against tuned cosine, WSD, and constant, with median relative differences of
$-0.28\%$, $-0.24\%$, and $-0.90\%$. None of these win rates differs
significantly from $50\%$. Per-task tuning matters here as the best peak varies by roughly a factor of five across the baselines. To quantify how much of that tuning the learned scheduler makes unnecessary, we take the draws where tuning lowers test loss by at least $0.5\%$ over the default and measure the fraction of that gain the learned scheduler achieves in one run, $(L_{\mathrm{default}}-L_{\mathrm{learned}})/(L_{\mathrm{default}}-L_{\mathrm{tuned}})$. The median recovery is $80\%$, $83\%$ and $78\%$ against cosine, WSD and constant (Figure~\ref{fig:h2-recovery}).

\begin{figure}[t]
  \centering
  \includegraphics[width=\linewidth]{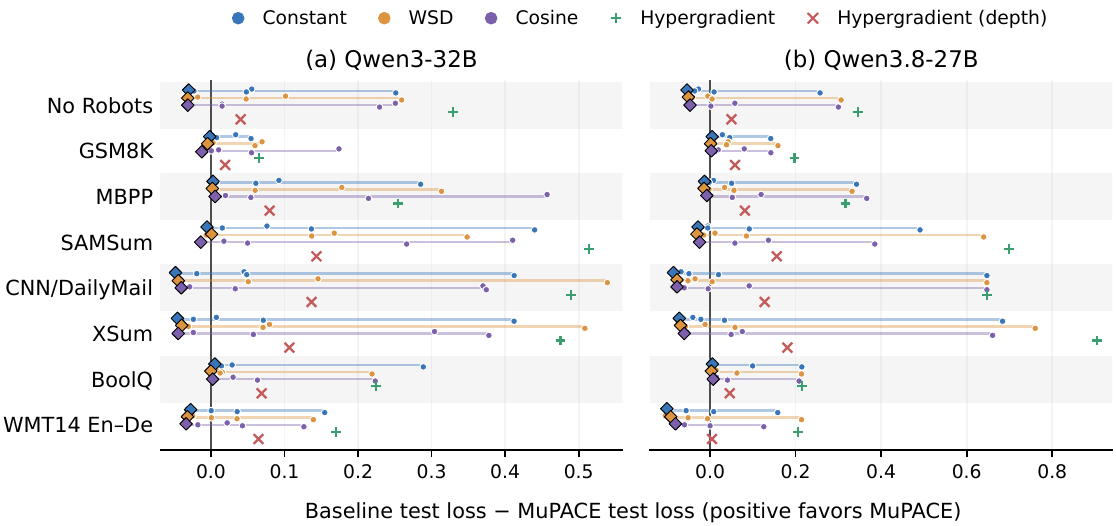}
  \vspace{-10pt}
  \caption{MuPACE versus validation-tuned baselines and Hypergradient Descent on \textbf{(a)} Qwen3-32B and \textbf{(b)} Qwen3.8-27B. Values show baseline minus MuPACE test loss (positive favors MuPACE).
}
  \label{fig:h3-comparison}
\end{figure}

\paragraph{H3: Meta-learned schedulers can transfer one-shot to out-of-distribution fine-tuning problems, avoiding per-task learning rate search while retaining competitive performance.} \label{H3}\mbox{}\\
Figure~\ref{fig:h3-comparison} evaluates MuPACE on models and datasets that were both unseen during policy training, comparing a single MuPACE run against fixed schedules with the peak tuned by a five-trial sweep on each dataset. The tuned baselines reach slightly lower test loss, with median relative differences of $1.13\%$--$2.43\%$ on Qwen3-32B and $2.59\%$--$3.02\%$ on Qwen3.8-27B, and MuPACE wins on $2$ of $8$ datasets per model. This gap is the price of one run versus five. Given only one trial, the fixed schedules perform worse than MuPACE by $14\%$--$18\%$ on average and need two to three trials to reach the same performance (Appendix~\ref{app:tuning-budget}). Against Hypergradient Descent, the other sweep-free method, MuPACE achieves lower test loss on every dataset for both models, whether Hypergradient Descent is used with or without $\mu$P post-scaling.

\paragraph{Design Ablations}
We ablate MuPACE's meta-training, schedule parameterization, shape normalization, and depth-dependent scaling to assess the contribution of each module to its performance.

\textit{Effect of meta-training.}
We compare against the same scheduler architecture with no meta-training, fixing all output heads at zero so it follows the default cosine schedule from \Cref{fig:schedule-family-parameters}. Meta-training lowers median in-distribution test loss and improves performance on Qwen3-32B (Appendix~\ref{app:untrained}).

\textit{Analytic schedule parameterization.}
We ablate the analytic schedule family by replacing the five heads defining~\Cref{eq:sched} with a single head that predicts the learning rate directly. Direct prediction has similar typical in-distribution performance but a heavier tail of poor runs and no wins against the baselines in the OOD setting, suggesting that the analytic parameterization primarily improves robustness (Appendix~\ref{app:direct-head}).

\textit{Shape normalization.}
We remove the shape normalization of \Cref{eq:norm}, so the shape
parameters can also change the schedule's overall magnitude. This has no
in-distribution effect, and on Qwen3-32B normalization is better on five of
eight datasets, a possible but not significant transfer benefit (Appendix~\ref{app:shape-norm}).

\textit{Depth-scaled baseline (fixed) schedules.}
We apply the same depth-dependent scaling \Cref{eq:depth} to each fixed-schedule baseline at (i) its default peak and (ii) its best peak from an in-distribution sweep. On both models, MuPACE achieves lower test loss in nearly all comparisons. These results suggest that MuPACE improves beyond transferring depth-scaled schedules, with in-distribution tuned peaks (Appendix~\ref{app:depth-controls}).

\textit{Depth-dependent post-scaling.}
We meta-train a variant without depth-dependent post-scaling. In-distribution performance is similar, but scaling improves test loss on all eight Qwen3-32B datasets, suggesting a benefit for one-shot transfer to larger models (Appendix~\ref{app:depth-post-scaling}).



\section{Conclusion}
\label{sec:conclusion}
We introduce MuPACE, a recurrent learning rate scheduler that adapts an interpretable analytic schedule to the observed optimization trajectory while leaving Muon's update rule unchanged. Across 64 fine-tuning problems, it outperforms strong default schedules and recovers 78--83\% of the benefit of per-task tuning in a single training run. Although meta-trained only on models with up to 16 layers, it transfers one-shot to 64-layer models and unseen tasks, where a single run remains competitive with swept schedules. These results suggest that amortizing learning-rate selection can remove per-task learning rate search while retaining competitive performance. 


\subsection*{AI use statement}
In this work, generative AI tools were used to assist with code implementation, support the authors’ understanding of relevant concepts, improve the clarity and readability of author-written text, and refine the visual presentation of plots. Generative AI tools were not used to formulate the original research question. Most of the related literature was identified independently by the authors, with AI tools used to supplement this process by suggesting additional relevant work.
All AI-assisted outputs were reviewed and verified by the authors. AI-assisted code was checked for correctness by the authors, literature-related suggestions were verified against the original sources, and all text and visual edits were reviewed before inclusion. The authors take full responsibility for the final content, claims, results, and artifacts presented in this work.

\subsection*{Reproducibility statement}

We open source our code for meta-training and evaluation, together with the meta-trained MuPACE checkpoint, at \url{https://anonymous.4open.science/r/MUPACE-3F54/README.md}. The out-of-distribution base models (Qwen3-32B \citep{yang2025qwen3} and Qwen3.8-27B \citep{qwen38}) are publicly available, and the nanochat models  used for meta-training can be retrained with the open-source nanochat implementation \citep{nanochat}. All datasets are publicly available (\citep{wang-etal-2022-super}; \Cref{app:ood_datasets}). \Cref{sec:method} fully specifies the method, including the analytic schedule family (Equations~\ref{eq:sched}--\ref{eq:depth}), the policy and its input features (\Cref{sec:meta-model}), and the meta-training objective and gradient estimator (\Cref{sec:outer}). \Cref{sec:protocol} describes the experimental protocol, including the meta-training setup and compute budget (\Cref{sec:arch-training}) and the evaluation and checkpoint-selection procedure (\Cref{sec:benchmarks}). \Cref{app:prior} lists the complete meta-training distribution, including all model configurations (\Cref{tab:model-deck}) and dataset identifiers (\Cref{tab:training-datasets}), and \Cref{app:hyperparams} reports the remaining hyperparameters.


\subsubsection*{Acknowledgments}
Josif Grabocka acknowledges the funding support from the "Bayerisches Landesamt für Steuern" for the Bavarian AI Taxation Laboratory.

Josif Grabocka acknowledges the funding by The Carl Zeiss Foundation through the research network "Responsive and Scalable Learning for Robots Assisting Humans" (ReScaLe) of the University of Freiburg.

Josif Grabocka acknowledges the financial support from the Albanian-American Development Foundation through the READ program.

Moreover, we gratefully acknowledge the scientific support and HPC resources provided by the Erlangen National High Performance Computing Center (NHR@FAU) of the Friedrich-Alexander-Universität Erlangen-Nürnberg (FAU). NHR@FAU hardware is partially funded by the German Research Foundation (DFG) – 440719683.

\bibliography{iclr2027_conference}
\bibliographystyle{iclr2027_conference}

\clearpage
\appendix
\raggedbottom

\let\origsection\section
\renewcommand\section{\FloatBarrier\origsection}

\section{Task Distribution}
\label{app:prior}
The meta-training distribution combines pretrained nanochat models with $100$ English Lots-of-LoRAs sub-tasks, including classification, question answering, paraphrasing, reasoning, summarization, and text generation. Each episode fine-tunes LoRA adapters on the Q and V projections of every attention block, with the base model frozen. Table~\ref{tab:task-prior} summarizes the meta-training configuration.

\begin{table}[ht]
  \centering
  
  \small
    \caption{Meta-training task distribution. Scale augmentation is disabled at evaluation.}
  \begin{tabular}{ll}
    \hline
    Component & Values or sampling range \\
    \hline
    Base-model deck & $37$ nanochat checkpoints \\
    Model depth / width & $\{8,12,16\}$ / $128$--$2048$ \\
    Dataset & $100$ Lots-of-LoRAs sub-tasks \\
    LoRA rank $r$ & $\{8,16,32\}$ \\
    LoRA scaling $a$ & $\mathrm{LogUniform}[8,32]$ \\
    Batch size $b$ & $\{8,16,32\}$ \\
    Episode budget $T$ & $\mathrm{LogUniform}[100,300]$ (integer steps) \\
    Sequence length & $128$ tokens \\
    Global scale augmentation $\kappa$ & $\mathrm{LogUniform}[0.05,20]$ \\
    \hline
  \end{tabular}
  \label{tab:task-prior}
\end{table}

\paragraph{Model list.}
The $37$ checkpoint identifiers are listed below, grouped by depth; in \texttt{d$L$\_w$W$}, $L$ denotes the number of layers and $W$ the hidden width.
\begin{table}[ht]
  \centering
  \small
\caption{Model list used during Meta-training}
\label{tab:model-deck}
\begin{tabular}{@{}p{3.4em}@{\hspace{1em}}p{\dimexpr\linewidth-4.4em\relax}@{}}
\hline
Depth & Checkpoint identifiers \\
\hline
8 & \texttt{d8\_w128}, \texttt{d8\_w256}, \texttt{d8\_w384}, \texttt{d8\_w512}, \texttt{d8\_w640}, \texttt{d8\_w768}, \texttt{d8\_w896}, \texttt{d8\_w1024}, \texttt{d8\_w1152}, \texttt{d8\_w1280}, \texttt{d8\_w1408}, \texttt{d8\_w1536}, \texttt{d8\_w1664}, \texttt{d8\_w1792}, \texttt{d8\_w1920}, \texttt{d8\_w2048} \\
12 & \texttt{d12\_w128}, \texttt{d12\_w256}, \texttt{d12\_w384}, \texttt{d12\_w512}, \texttt{d12\_w640}, \texttt{d12\_w768}, \texttt{d12\_w896}, \texttt{d12\_w1024}, \texttt{d12\_w1152}, \texttt{d12\_w1280}, \texttt{d12\_w1408}, \texttt{d12\_w1664} \\
16 & \texttt{d16\_w128}, \texttt{d16\_w256}, \texttt{d16\_w384}, \texttt{d16\_w512}, \texttt{d16\_w640}, \texttt{d16\_w896}, \texttt{d16\_w1024}, \texttt{d16\_w1280}, \texttt{d16\_w1408} \\
\hline
\end{tabular}
\end{table}

\paragraph{Dataset list.}
The following $100$ identifiers appear in launch-script order. Each is prefixed
with \texttt{Lots-of-LoRAs/} to obtain the full dataset name \citep{wang-etal-2022-super}.
\begingroup
\footnotesize
\setlength{\tabcolsep}{3pt}
\setlength{\LTleft}{\fill}
\setlength{\LTright}{\fill}
\begin{longtable}{@{}p{2.4em}@{\hspace{1em}}p{\dimexpr\linewidth-3.4em\relax}@{}}
\caption{Lots-of-LoRAs sub-tasks used for meta-training.}\label{tab:training-datasets}\\
\hline
No. & Dataset identifier (without organization prefix) \\
\hline
\endfirsthead
\hline
No. & Dataset identifier (continued) \\
\hline
\endhead
\hline
\endfoot
1 & \nolinkurl{task045_miscellaneous_sentence_paraphrasing} \\
2 & \nolinkurl{task588_amazonfood_rating_classification} \\
3 & \nolinkurl{task461_qasper_question_generation} \\
4 & \nolinkurl{task083_babi_t1_single_supporting_fact_answer_generation} \\
5 & \nolinkurl{task280_stereoset_classification_stereotype_type} \\
6 & \nolinkurl{task1332_check_leap_year} \\
7 & \nolinkurl{task296_storycloze_correct_end_classification} \\
8 & \nolinkurl{task209_stancedetection_classification} \\
9 & \nolinkurl{task400_paws_paraphrase_classification} \\
10 & \nolinkurl{task316_crows-pairs_classification_stereotype} \\
11 & \nolinkurl{task1565_triviaqa_classification} \\
12 & \nolinkurl{task1592_yahoo_answers_topics_classfication} \\
13 & \nolinkurl{task284_imdb_classification} \\
14 & \nolinkurl{task137_detoxifying-lms_classification_toxicity} \\
15 & \nolinkurl{task1645_medical_question_pair_dataset_text_classification} \\
16 & \nolinkurl{task328_jigsaw_classification_insult} \\
17 & \nolinkurl{task1286_openbookqa_question_answering} \\
18 & \nolinkurl{task703_mmmlu_answer_generation_high_school_geography} \\
19 & \nolinkurl{task686_mmmlu_answer_generation_college_biology} \\
20 & \nolinkurl{task648_answer_generation} \\
21 & \nolinkurl{task492_mwsc_incorrect_answer_generation} \\
22 & \nolinkurl{task1598_nyc_long_text_generation} \\
23 & \nolinkurl{task593_sciq_explanation_generation} \\
24 & \nolinkurl{task110_logic2text_sentence_generation} \\
25 & \nolinkurl{task183_rhyme_generation} \\
26 & \nolinkurl{task294_storycommonsense_motiv_text_generation} \\
27 & \nolinkurl{task568_circa_question_generation} \\
28 & \nolinkurl{task1729_personachat_generate_next} \\
29 & \nolinkurl{task275_enhanced_wsc_paraphrase_generation} \\
30 & \nolinkurl{task770_pawsx_english_text_modification} \\
31 & \nolinkurl{task116_com2sense_commonsense_reasoning} \\
32 & \nolinkurl{task1421_mathqa_other} \\
33 & \nolinkurl{task632_dbpedia_14_classification} \\
34 & \nolinkurl{task746_yelp_restaurant_review_classification} \\
35 & \nolinkurl{task846_pubmedqa_classification} \\
36 & \nolinkurl{task297_storycloze_incorrect_end_classification} \\
37 & \nolinkurl{task1705_ljspeech_classification} \\
38 & \nolinkurl{task1531_daily_dialog_type_classification} \\
39 & \nolinkurl{task1721_civil_comments_obscenity_classification} \\
40 & \nolinkurl{task1575_amazon_reviews_multi_sentiment_classification} \\
41 & \nolinkurl{task1488_sarcasmdetection_headline_classification} \\
42 & \nolinkurl{task379_agnews_topic_classification} \\
43 & \nolinkurl{task322_jigsaw_classification_threat} \\
44 & \nolinkurl{task1495_adverse_drug_event_classification} \\
45 & \nolinkurl{task108_contextualabusedetection_classification} \\
46 & \nolinkurl{task472_haspart_classification} \\
47 & \nolinkurl{task521_trivia_question_classification} \\
48 & \nolinkurl{task495_semeval_headline_classification} \\
49 & \nolinkurl{task1593_yahoo_answers_topics_classification} \\
50 & \nolinkurl{task1216_atomic_classification_causes} \\
51 & \nolinkurl{task1722_civil_comments_threat_classification} \\
52 & \nolinkurl{task838_cdt_classification} \\
53 & \nolinkurl{task081_piqa_wrong_answer_generation} \\
54 & \nolinkurl{task667_mmmlu_answer_generation_business_ethics} \\
55 & \nolinkurl{task1327_qa_zre_answer_generation_from_question} \\
56 & \nolinkurl{task344_hybridqa_answer_generation} \\
57 & \nolinkurl{task1581_eqasc-perturbed_answer_generation} \\
58 & \nolinkurl{task471_haspart_answer_generation} \\
59 & \nolinkurl{task307_jeopardy_answer_generation_final} \\
60 & \nolinkurl{task073_commonsenseqa_answer_generation} \\
61 & \nolinkurl{task727_mmmlu_answer_generation_prehistory} \\
62 & \nolinkurl{task1549_wiqa_answer_generation_missing_step} \\
63 & \nolinkurl{task033_winogrande_answer_generation} \\
64 & \nolinkurl{task1557_jfleg_answer_generation} \\
65 & \nolinkurl{task1326_qa_zre_question_generation_from_answer} \\
66 & \nolinkurl{task074_squad1.1_question_generation} \\
67 & \nolinkurl{task1325_qa_zre_question_generation_on_subject_relation} \\
68 & \nolinkurl{task167_strategyqa_question_generation} \\
69 & \nolinkurl{task442_com_qa_paraphrase_question_generation} \\
70 & \nolinkurl{task897_freebase_qa_topic_question_generation} \\
71 & \nolinkurl{task754_svamp_common-division_question_answering} \\
72 & \nolinkurl{task753_svamp_addition_question_answering} \\
73 & \nolinkurl{task868_mawps_singleop_question_answering} \\
74 & \nolinkurl{task866_mawps_multidiv_question_answering} \\
75 & \nolinkurl{task671_ambigqa_text_generation} \\
76 & \nolinkurl{task131_scan_long_text_generation_action_command_long} \\
77 & \nolinkurl{task1711_poki_text_generation} \\
78 & \nolinkurl{task128_scan_structured_text_generation_command_action_short} \\
79 & \nolinkurl{task210_logic2text_structured_text_generation} \\
80 & \nolinkurl{task899_freebase_qa_topic_generation} \\
81 & \nolinkurl{task1582_bless_hypernym_generation} \\
82 & \nolinkurl{task300_storycloze_order_generation} \\
83 & \nolinkurl{task102_commongen_sentence_generation} \\
84 & \nolinkurl{task926_coached_conv_pref_word_generation} \\
85 & \nolinkurl{task675_google_wellformed_query_sentence_generation} \\
86 & \nolinkurl{task828_copa_commonsense_cause_effect} \\
87 & \nolinkurl{task291_semeval_2020_task4_commonsense_validation} \\
88 & \nolinkurl{task1152_bard_analogical_reasoning_causation} \\
89 & \nolinkurl{task1507_boolean_temporal_reasoning} \\
90 & \nolinkurl{task181_outcome_extraction} \\
91 & \nolinkurl{task1510_evalution_relation_extraction} \\
92 & \nolinkurl{task1452_location_entity_extraction_btc_corpus} \\
93 & \nolinkurl{task1562_zest_text_modification} \\
94 & \nolinkurl{task121_zest_text_modification} \\
95 & \nolinkurl{task645_summarization} \\
96 & \nolinkurl{task288_gigaword_summarization} \\
97 & \nolinkurl{task1285_kpa_keypoint_matching} \\
98 & \nolinkurl{task755_find_longest_substring_and_replace_its_sorted_lowercase_version_in_both_lists} \\
99 & \nolinkurl{task206_collatz_conjecture} \\
100 & \nolinkurl{task1502_hatexplain_classification} \\
\end{longtable}
\endgroup

\subsection{OOD Evaluation Datasets}\label{app:ood_datasets}
\Cref{tab:ood_datasets} lists the eight publicly available datasets used for out-of-distribution evaluation, none of which are among the 100 meta-training sub-tasks from \Cref{tab:training-datasets}.

\begin{table}[h]
\centering
\caption{Out-of-distribution evaluation datasets.}
\label{tab:ood_datasets}
\begin{tabular}{l|l}
\hline
Dataset & Domain  \\
\hline
No Robots \citep{no_robots}        & Instruction following \\
GSM8K \citep{cobbe2021gsm8k}            & Mathematical reasoning \\
MBPP \citep{austin2021program}             & Code generation \\
SAMSum \citep{gliwa-etal-2019-samsum}           & Dialogue summarization \\
CNN/DailyMail \citep{see-etal-2017-get}    & News summarization \\
XSum \citep{Narayan2018DontGM}             & Extreme (single-sentence) summarization \\
BoolQ \citep{clark2019boolq}            & Question answering (yes/no) \\
WMT14 En$\rightarrow$De \citep{bojar-EtAl:2014:W14-33} & Machine translation \\
\hline
\end{tabular}
\end{table}

\section{Hyperparameters}
\label{app:hyperparams}
Tables~\ref{tab:training-hyperparams} and~\ref{tab:schedule-hyperparams} summarize the settings used for meta-training. The task distribution is specified in Appendix~\ref{app:prior}.

\begin{table}[H]
  \centering
  \small
  \caption{Policy architecture and meta-training hyperparameters.}
  \label{tab:training-hyperparams}
  \begin{tabular}{@{}ll@{}}
    \hline
    Hyperparameter & Value \\
    \hline
    Input features / LSTM hidden width & $20$ / $48$ \\
    Meta-parameters & $19{,}877$ \\
    Output head initialization  & Zero \\
    Outer optimizer & AdamW, weight decay $10^{-4}$ \\
    Outer learning rate & $10^{-3}$, cosine decay \\
    Meta-gradient clipping & Global norm $1$ \\
    Outer iterations  & $9000$ \\
    PES perturbation standard deviation & $0.01$ \\
    PES truncation / parallel task instances & $50$ steps / $8$ \\
    Objective start fraction $f$ & $0.2$ (last $80\%$ of each episode) \\
    Fast / slow loss-feature EMA decays & $0.9$ / $0.97$ \\
    Muon momentum / Newton--Schulz iterations & $0.95$ / $5$ \\
    Depth-scaling exponent / reference depth & $0.5$ / $11.3137$ \\
    \hline
  \end{tabular}
\end{table}

\begin{table}[H]
  \centering
  \small
  \caption{Schedule defaults and adjustment ranges. Each head emits
  $\lambda_i=\lambda_i^0\exp(c_i\tanh h_i)$, giving the range
  $[\lambda_i^0 e^{-c_i},\lambda_i^0 e^{c_i}]$.}
  \label{tab:schedule-hyperparams}
  \begin{tabular}{@{}lcc@{}}
    \hline
    Parameter & Default $\lambda_i^0$ & Reach $c_i$ \\
    \hline
    Level $\lambda_1$ & $10^{-3}$ & $4.5$ \\
    Decay exponent $\lambda_2$ & $2$ & $1.0$ \\
    Time-warp exponent $\lambda_3$ & $1$ & $1.4$ \\
    Tail floor $\lambda_4$ & $0.01$ & $2.3$ \\
    Warmup time constant $\lambda_5$ & $0.03$ & $0.7$ \\
    \hline
  \end{tabular}
\end{table}

The warmup parameter is fixed after the first policy evaluation of each episode.
Shape normalization uses $65$ uniformly spaced points on $[0,1]$, with cosine clamp $\delta=10^{-4}$. The normalized shape is capped at three times the default shape's peak before applying the level and depth factors.

\clearpage
\section{Tuning Budget and MuPACE Loss Curves}
\label{app:tuning-budget}

\begin{figure}[H]
  \centering
  \includegraphics[width=\linewidth]{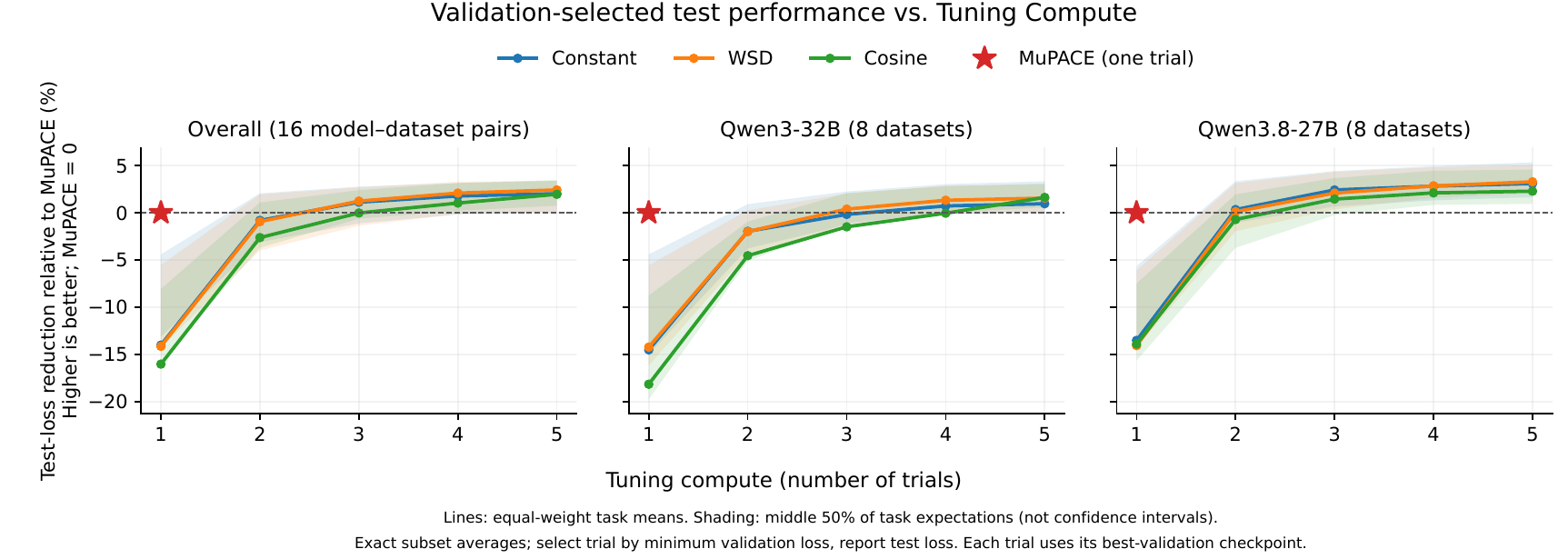}
  \caption{Expected test performance under validation-based trial selection. Shading shows the interquartile range across task expectations, not confidence intervals.}
  \label{fig:trial-budget}
\end{figure}

Figure~\ref{fig:trial-budget} shows how baseline performance changes when we allow more learning-rate trials, while MuPACE uses only one. For each budget from one to five trials, we consider every possible group of that size. In each group, we select the trial with the lowest validation loss and record its test loss. We then average these results. The plot shows the percentage reduction in test loss relative to MuPACE, so positive values favor the baseline. Each dataset has equal weight, and the overall panel combines both models.

\begin{figure}[H]
  \centering
  \includegraphics[width=0.88\linewidth]{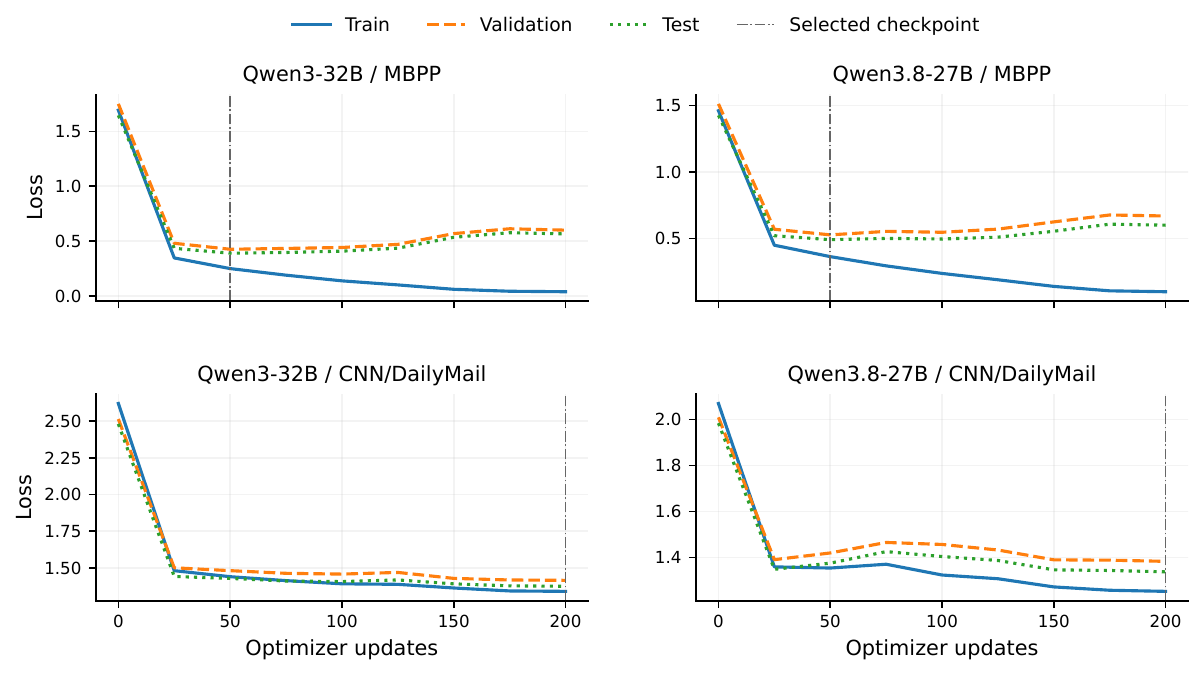}
  \caption{Illustrative MuPACE train, validation, and test evaluation losses on MBPP and CNN/DailyMail.}
  \label{fig:mupace-loss-curves}
\end{figure}

\section{Ablation: Untrained Scheduler}
\label{app:untrained}

We compare MuPACE against its untrained counterpart, whose output head is fixed to zero, corresponding to its initialization before meta-learning. This variant therefore produces the default cosine schedule shown in \Cref{fig:schedule-family-parameters}, together with depth-dependent post-scaling. The comparison isolates the contribution of meta-training beyond the schedule initialization.

\begin{figure}[H]
  \centering
  \includegraphics[width=0.8\linewidth]{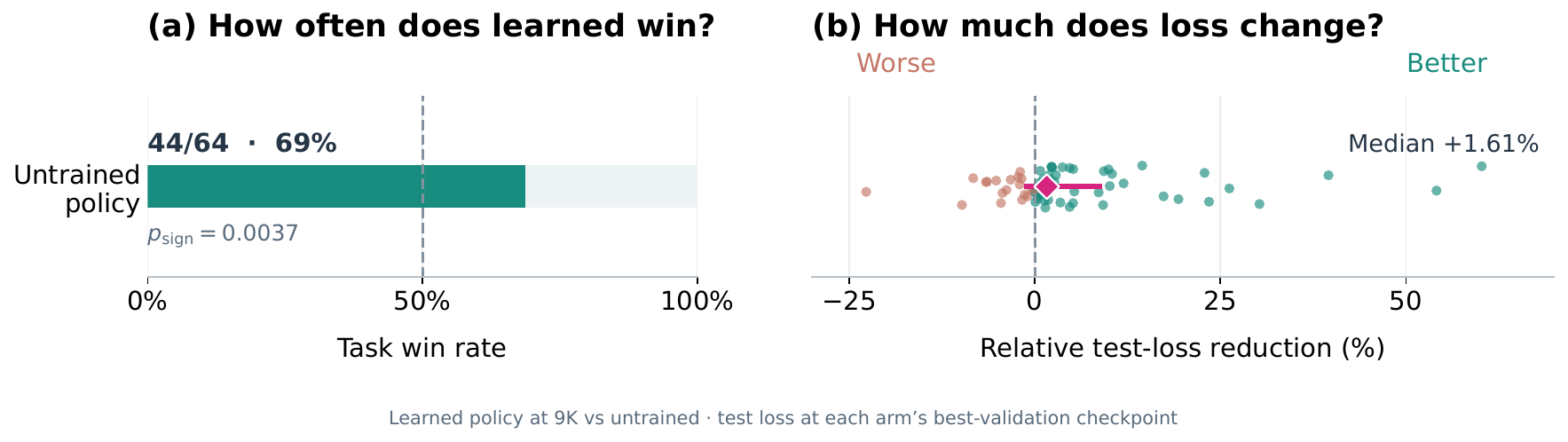}\\[1ex]
  \includegraphics[width=0.49\linewidth]{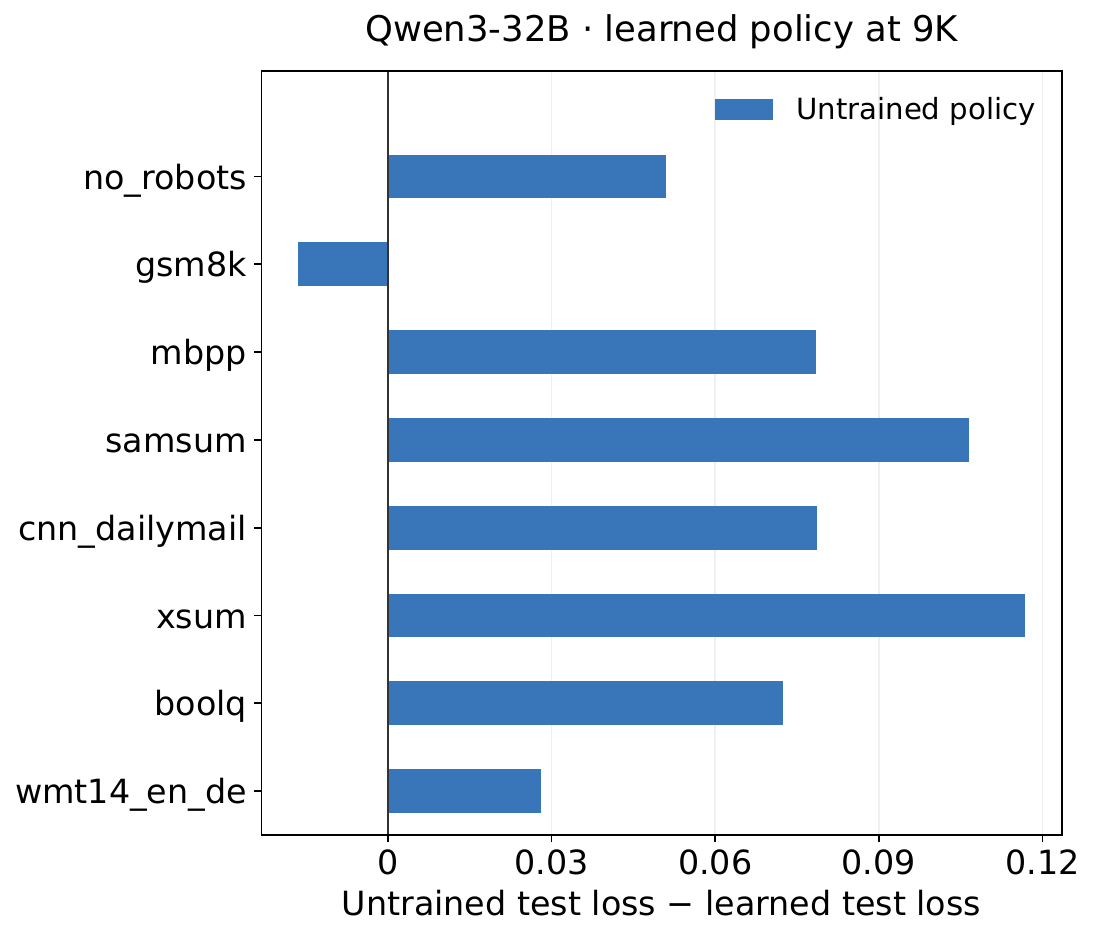}
  \caption{MuPACE versus the untrained zero-head policy on in-distribution draws (top) and Qwen3-32B (bottom).}
  \label{fig:abl-untrained}
\end{figure}

On the $64$ in-distribution draws, our meta-learned MuPACE wins on $44$ and reduces median test loss by $1.61\%$ (two-sided exact sign test $p=0.0037$). On Qwen3-32B, it wins on $7/8$ datasets, with a median reduction of $6.66\%$ (sign test $p=0.0703$). Figure~\ref{fig:abl-untrained} shows the comparisons.

\section{Ablation: Direct Learning-Rate Head}
\label{app:direct-head}

\begin{figure}[H]
  \centering
  \includegraphics[width=0.48\linewidth]{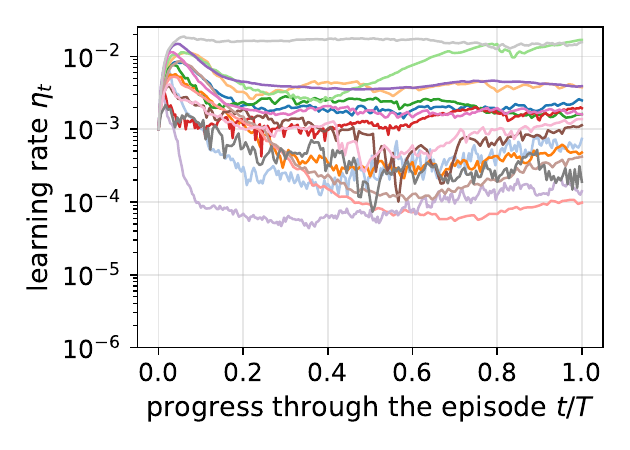}
  \caption{Learning rates applied by the direct LR head on $16$ in-distribution tasks.}
  \label{fig:learned-schedules-lr}
\end{figure}

In the following ablation, we replace the five schedule heads of \Cref{eq:sched} with a single head that directly predicts a learning rate before depth-$\mu$P post-scaling. We extend its range downward (approximately $[1.1\times10^{-8},\,9.0\times10^{-2}]$) to accommodate the small learning rates produced by the normalized analytic schedule (especially $\lambda_4$). Everything else remains unchanged. At initialization, this head outputs a constant $10^{-3}$ before post-scaling, so it must learn how the rate changes over time, including warm-up. The policy is meta-trained for $9000$ outer iterations. Figure~\ref{fig:learned-schedules-lr} shows example schedules after meta-training, demonstrating that warm-up emerges through learning rather than being set by the initialization.

\begin{figure}[!htbp]
  \centering
  \includegraphics[width=0.8\linewidth]{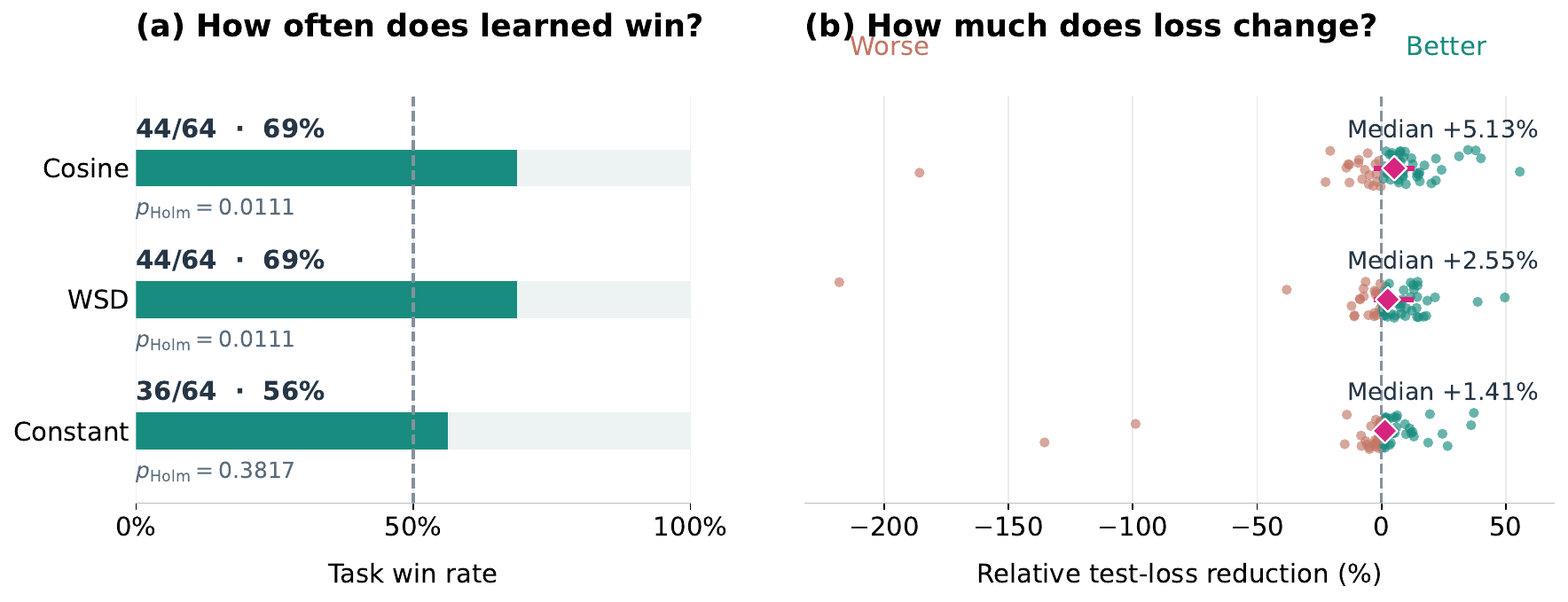}\\
  \small (a) versus default schedules\\[1ex]
  \includegraphics[width=0.8\linewidth]{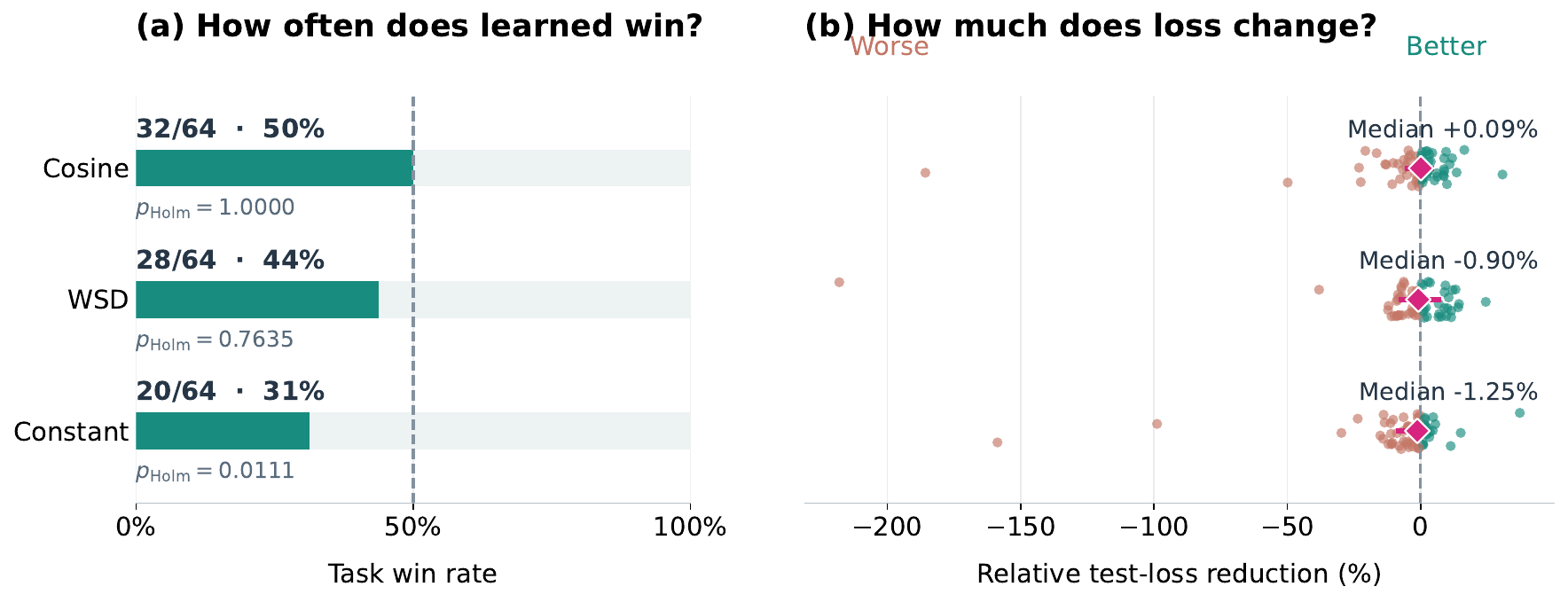}\\
  \small (b) versus tuned schedules
  \caption{Direct learning-rate head on the same $64$ in-distribution draws. \textbf{(a)} compares against the default schedules, while \textbf{(b)} compares against the swept baselines.}
  \label{fig:abl-direct}
\end{figure}

\begin{figure}[!htbp]
  \centering
  \includegraphics[width=0.53\linewidth]{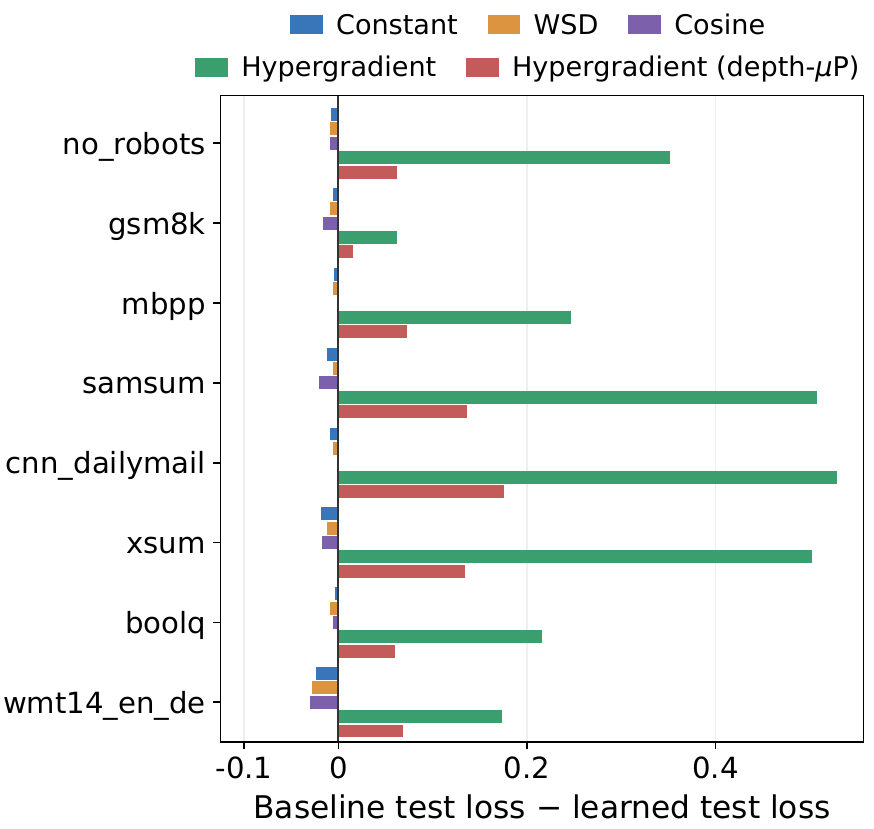}
  \caption{Direct learning-rate head transferred one-shot to Qwen3-32B. Bars show baseline minus learned test loss, where positive values favor the learned scheduler.
}
  \label{fig:abl-direct-qwen}
\end{figure}

The direct head clearly outperforms default cosine and WSD, but not default constant, and performs worse than tuned constant (Figure~\ref{fig:abl-direct}). Its median performance is close to that of the analytic scheduler, with a difference of only $0.4\%$ that is not significant ($p=0.38$), although it performs worse on a few tasks. On Qwen3-32B, it trails all tuned baselines across the eight datasets but consistently outperforms Hypergradient Descent (Figure~\ref{fig:abl-direct-qwen}). Overall, the analytic parameterization appears to improve robustness more than typical performance.

\section{Ablation: Shape Normalization}
\label{app:shape-norm}

In the following ablation, we remove the shape normalization of Equation~\ref{eq:norm}, so the shape parameters can also change the overall magnitude of the schedule. All other components remain unchanged, and both policies are meta-trained for $9000$ outer iterations.

\begin{figure}[H]
  \centering
  \includegraphics[width=0.8\linewidth]{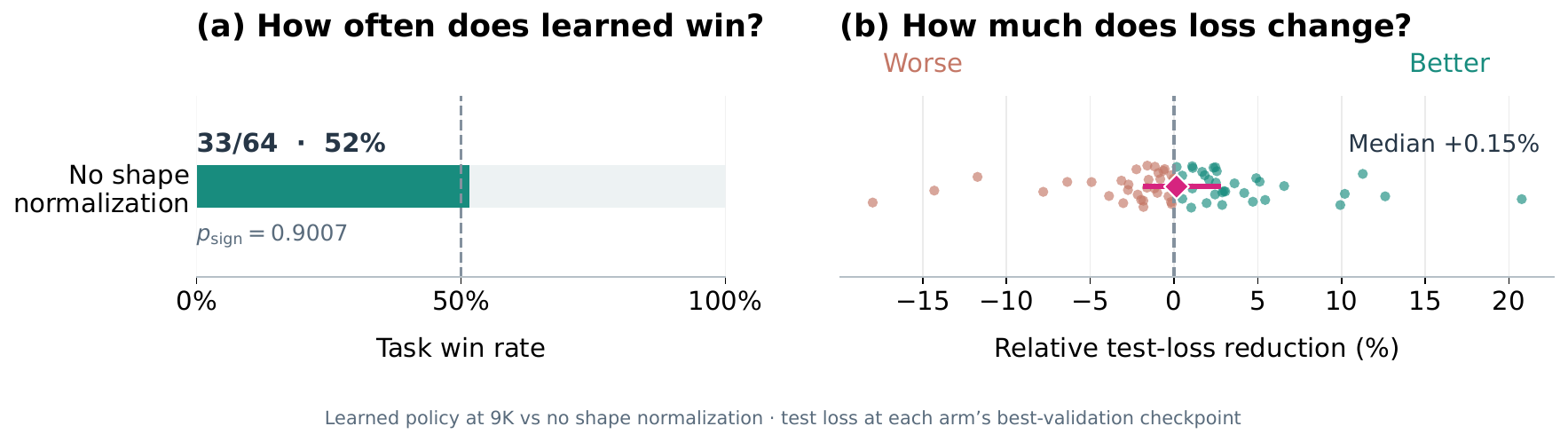}\\[1ex]
  \includegraphics[width=0.49\linewidth]{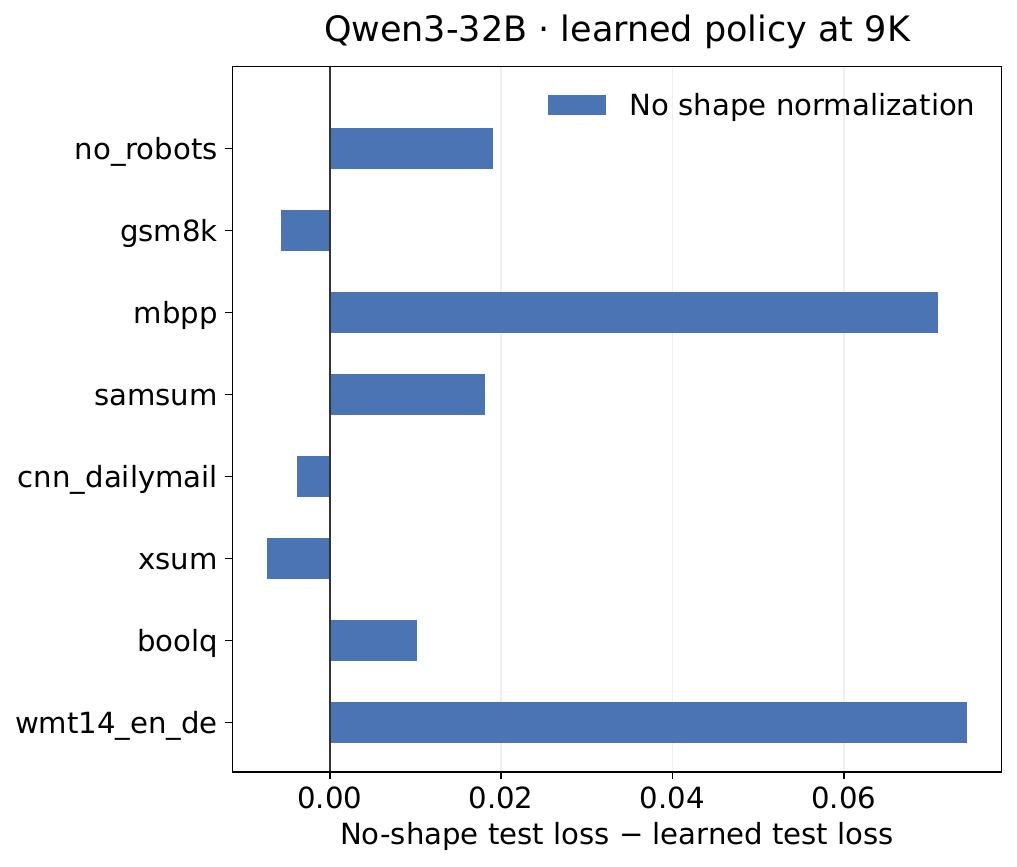}
  \caption{Comparison of MuPACE with and without shape normalization on in-distribution draws (top) and Qwen3-32B (bottom). Positive values mean shape normalization performs better.}
  \label{fig:abl-shape-norm}
\end{figure}

On the $64$ in-distribution draws, the normalized scheduler wins on $33$ and reduces median test loss by $0.15\%$ (two-sided exact sign test $p=0.90$), so we find no in-distribution effect. On Qwen3-32B, it wins on $5/8$ datasets, with a median reduction of $1.37\%$ (sign test $p=0.73$). The differences are small on most datasets, but large where normalization helps, most clearly on mbpp ($15.4\%$) and wmt14\_en\_de ($8.3\%$). With eight datasets, this suggests a possible transfer benefit rather than a significant one. Figure~\ref{fig:abl-shape-norm} shows the comparisons.

\section{Ablation: Depth Scaling}
\label{app:depth-controls}

\subsection{Depth-scaled Baseline fixed schedules}
We ask whether MuPACE still outperforms the fixed schedules when the latter are given the same depth-dependent scaling and their learning-rate peaks are selected without any target-task search. Specifically, we apply the learned policy's depth-$\mu$P multiplier from Equation~\ref{eq:depth} to cosine, WSD, and constant schedules on both out-of-distribution models. For each schedule family, we transfer (i) the default peak of $10^{-3}$ and (ii) the best fixed peak from a sweep on the in-distribution evaluation, denoted $p^*$. For WSD and constant, the source-selected and default peaks coincide. Thus, the comparison tests whether MuPACE provides gains beyond transferring either default or source-tuned depth-scaled fixed schedules. All arms report test loss at their own best-validation checkpoint.

\begin{figure}[H]
  \centering
  \includegraphics[width=\linewidth]{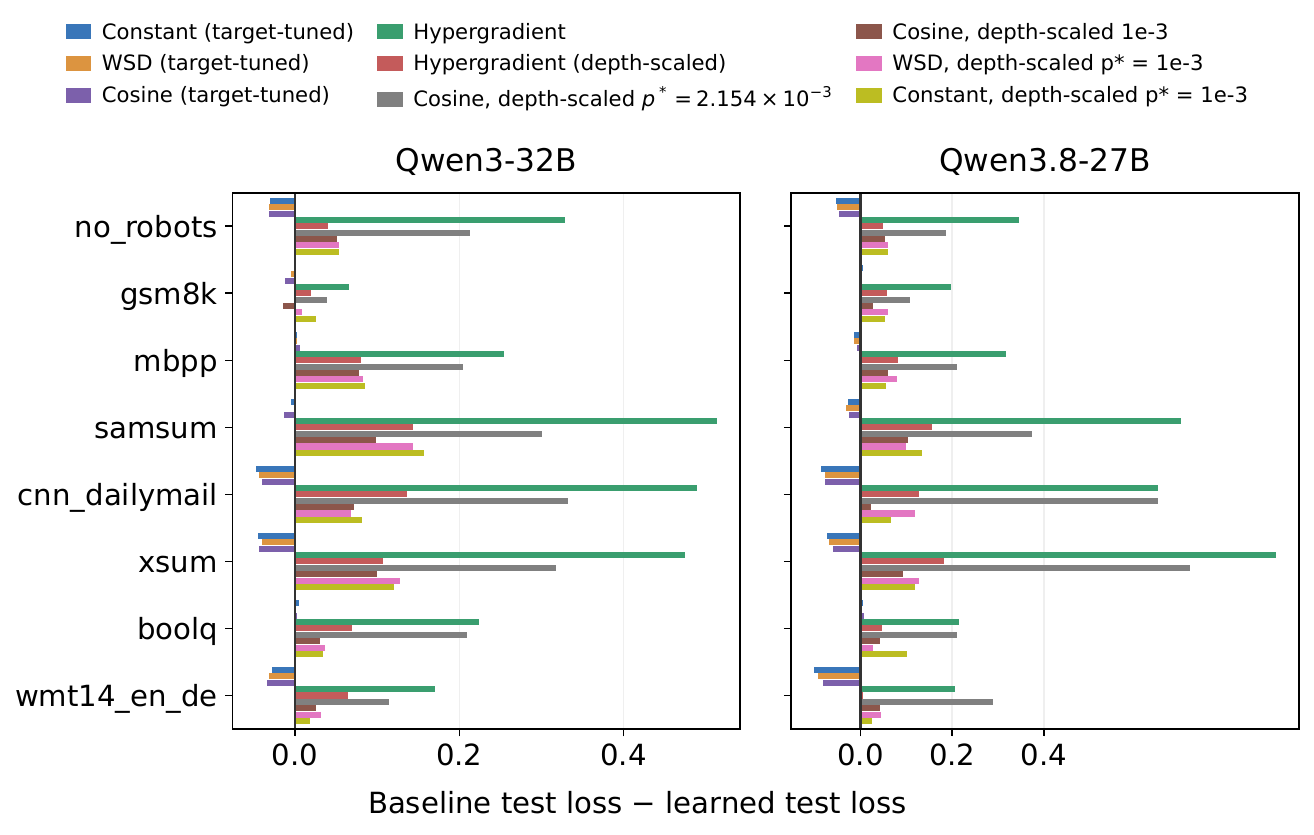}
  \caption{Depth-scaled fixed schedules versus MuPACE on Qwen3-32B (left) and Qwen3.8-27B (right). Positive differences favor the learned policy. Target-tuned schedules and Hypergradient Descent are shown for context. Depth-scaled fixed schedules use no target-task peak selection.}
  \label{fig:depth-controls}
\end{figure}

MuPACE outperforms every depth-scaled fixed control on every dataset of both models except default-peak cosine on GSM8K for Qwen3-32B (Figure~\ref{fig:depth-controls}). This supports gains beyond applying the same depth-scaling rule to default or source-tuned fixed schedules without target-task search.

\subsection{Depth-dependent post-scaling of the learned policy}
\label{app:depth-post-scaling}

In the following ablation, we meta-train a separate variant without the depth-dependent multiplier from Equation~\ref{eq:depth}, while retaining depth as a policy input. We compare the two policies on the same $64$ in-distribution draws and eight Qwen3-32B datasets, reporting test loss at each arm's best-validation checkpoint.

\begin{figure}[H]
  \centering
  \includegraphics[width=0.8\linewidth]{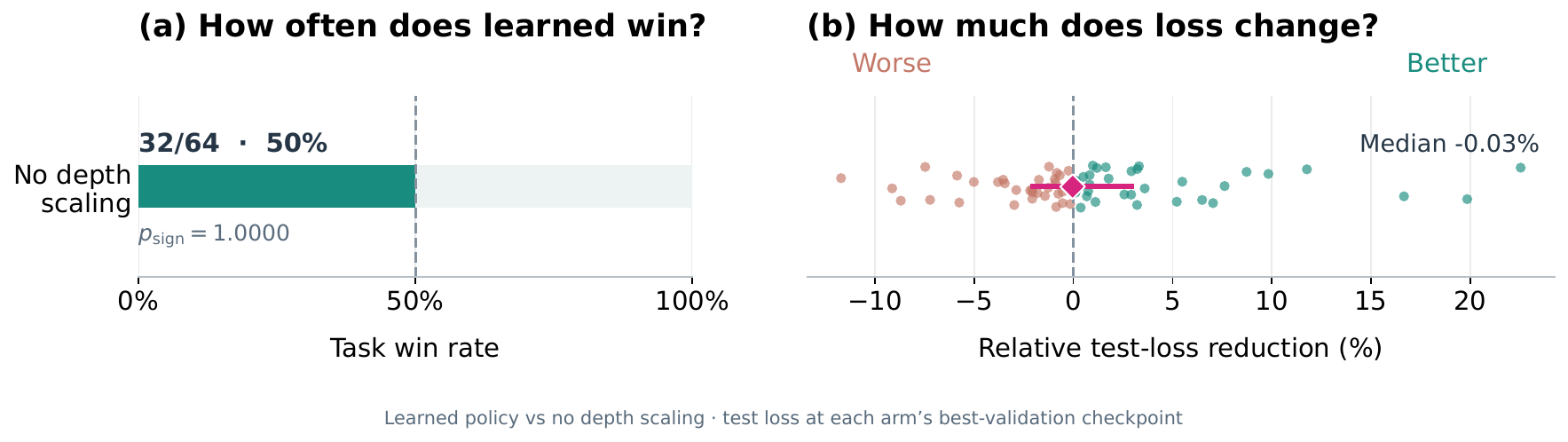}\\[1ex]
  \includegraphics[width=0.49\linewidth]{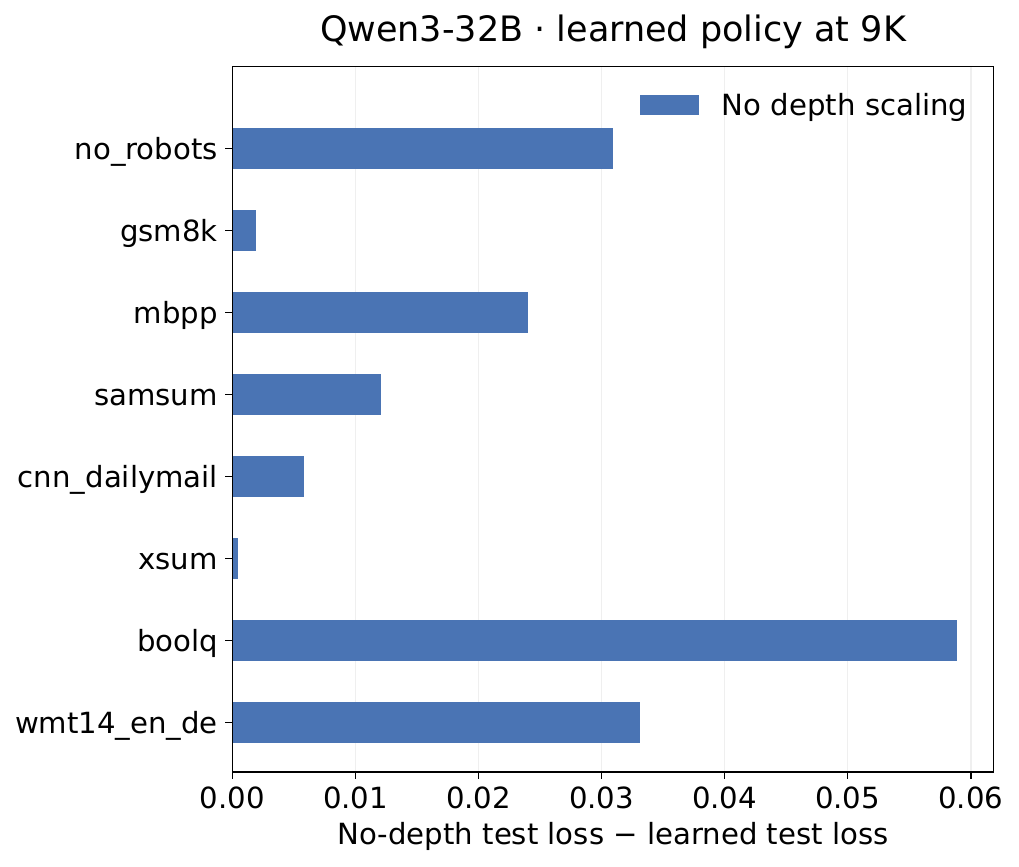}
  \caption{MuPACE versus a separately meta-trained variant without depth-dependent post-scaling on in-distribution draws (top) and Qwen3-32B datasets (bottom). Positive values favor the full scheduler.}
  \label{fig:abl-depth-post-scaling}
\end{figure}

In-distribution performance is similar, with MuPACE winning on $32/64$ draws and achieving a median relative test-loss reduction of $-0.03\%$ (two-sided exact sign test $p=1.00$). On Qwen3-32B, MuPACE wins on all eight datasets, suggesting a benefit for one-shot transfer (Figure~\ref{fig:abl-depth-post-scaling}). 

\end{document}